\documentclass{article}
\usepackage{spconf,amsmath,epsfig}
\usepackage{algorithm,algorithmic,cite,amsmath,amssymb,epsfig,enumerate,amsfonts,url,multirow,array,tabularx,booktabs,stfloats,flushend}
\usepackage[T1]{fontenc}

\def\b{{\mathbf b}}
\def\x{{\mathbf x}}

\def\z{{\mathbf z}}

\def\W{{\mathbf W}}

\title{Denoising Deep Neural Networks Based Voice Activity Detection}
\name{Xiao-Lei Zhang and Ji Wu}
\address{Multimedia Signal and Intelligent Information
Processing Laboratory, \\
Tsinghua National Laboratory for Information Science and Technology,\\
Department of Electronic Engineering,
Tsinghua University, Beijing, China.\\
huoshan6@126.com, wu\_ji@tsinghua.edu.cn
\thanks{This work is supported in part by the China Postdoctoral Science Foundation funded project under Grant 2012M520278 and in part by the National Natural Science Funds of China under Grant 61170197.}
}

\begin{document}
%
\maketitle
\begin{abstract}
Recently, the deep-belief-networks (DBN) based voice activity detection (VAD) has been proposed. It is powerful in fusing the advantages of multiple features, and achieves the state-of-the-art performance. However, the deep layers of the DBN-based VAD do not show an apparent superiority to the shallower layers. In this paper, we propose a denoising-deep-neural-network (DDNN) based VAD to address the aforementioned problem. Specifically, we pre-train a deep neural network in a special unsupervised denoising greedy layer-wise mode, and then fine-tune the whole network in a supervised way by the common back-propagation algorithm. In the pre-training phase, we take the noisy speech signals as the visible layer and try to extract a new feature that minimizes the reconstruction cross-entropy loss between the noisy speech signals and its corresponding clean speech signals. Experimental results show that the proposed DDNN-based VAD not only outperforms the DBN-based VAD but also shows an apparent performance improvement of the deep layers over shallower layers.

\end{abstract}
\begin{keywords}
Deep learning, denoising deep neural networks, voice activity detection.
\end{keywords}
\section{Introduction}
\label{sec:intro}
Voice activity detectors (VADs) help to separate speech from its background noises. They are important frontends of modern speech processing systems, such as speech recognition systems \cite{yu2010deep,dahl2012context,hinton2012deep} and speech communication systems \cite{hanaclassification}. Recently, the machine-learning-based VADs have received much attention in that they have the following notable merits. First, they can be integrated to the speech recognition systems naturally. Second, they have strong theoretical bases that guarantee the performance. Third, they can fuse the advantages of multiple features \cite{wang2005time,wang2006computational,wang2012exploring,wang2012cocktail,wang2013towards} much better than traditional VADs.

The machine-learning-based VADs can be categorized to four groups \cite{kang2008discriminative,shin2010voice,yu2010discriminative,wu2011mmc,wu2011efficient,zhang2012linearithmic,suh2012multiple}. The first group is the discriminative-weight-training-based VADs \cite{kang2008discriminative,yu2010discriminative,suh2012multiple}. They conduct linear weighted combinations of multiple features in the original feature space. The second group is the support-vector-machine (SVM) based VADs \cite{shin2010voice,wu2011mmc}. They first fuse multiple features to a long feature vector in the original feature space, and then project the long feature vector to the kernel-induced feature space for better classification performance.
The third group is the multiple-kernel-SVM (MK-SVM) based VAD \cite{wu2011efficient,zhang2012linearithmic}. It takes the distribution diversity of multiple features into consideration by first projecting different features into different kernel spaces and then fuse the features in the kernel spaces in a way with linear weighted combination. All of the aforementioned three groups utilize shallow models, i.e. models with only zero or one hidden layer, which lack the ability of describing highly variant features and discovering the underlying manifold of the features.

The fourth group is the deep-belief-networks (DBNs) based VAD \cite{zhang2013deep}. Fundamentally, because the DBN \cite{hinton2006reducing} contains multiple hidden layers, the DBN-based VAD can describe highly variant features; because the unsupervised pre-training phase of DBN provides an initial point that is close to a good solution, the DBN-based VAD has a strong generalization ability when compared with other machine-learning-based VADs. However, the deep layers of the DBN-based VAD do not yield an apparent superiority to the shallower layers. In our personal opinion, it might not be proper to simply consider VAD as a binary-class classification problem with the noisy speech and the background noise as the two classes, since the background noise also contributes to the distribution of the noisy speech. This might account for the inapparent superiority of the deep layers over shallower layers in the DBN-based VAD.

In this paper, we propose a novel denoising deep-neural-networks (DDNNs) based VAD. The DDNN training also consists of two phases. The first phase is a special unsupervised denoising greedy layer-wise pre-training phase. The pre-training process of each hidden layer tries to extract a new feature that minimizes the \textit{reconstruction cross-entropy loss} between the noisy speech signals and its corresponding clean speech signals (but not the noisy speech signals). The second phase is the well-known supervised fine-tuning phase. It groups all layers with the pre-trained parameters to a whole deep neural networks and tune the parameters for the minimum classification error. Experimental results show that the proposed DDNN-based VAD not only outperforms the DBN-based VAD but also shows an apparent performance improvement of the deep layers over shallower layers.

\section{Denoising-DNN-based VAD}\label{sec:svm}
The training process of the DDNN-based VAD consists of two phases -- unsupervised denoising layer-wise pre-training phase and supervised fine-tuning phase, which are presented in detail in Sections \ref{subsec:udpr} and \ref{subsec:fine-tune} respectively. The overview of the DDNN-based VAD is presented in Algorithm 1.
\subsection{Unsupervised Denoising Layer-wise Pre-training}\label{subsec:udpr}
Suppose we have $D$-dimensional noisy speech observations (i.e. frames) $\left\{\x_i,y_i \right\}_{i=1}^{n}$ and their corresponding clean speech observations $\left\{ \tilde{\x}_i,y_i \right\}_{i=1}^{n}$ with $\x_i = [x_{i,d}]_{d=1}^{D}$, $y_i\in\{H_0,H_1\}$, where $x_d\in[0,1]$ and $H_1/H_0$ denote the speech$/$noise hypothesis.

The layer-wise pre-training of each module of DDNN consists of optimizing two activation functions jointly. The first function, denoted as $f_{\theta}(\cdot)$, maps the noisy speech observation from the visible layer $\x$ to a hidden layer $f_{\theta}(\x)$. The second function, denoted as $g_{\theta'}(\cdot)$, tries to reconstruct $\tilde{\x}$ (but not ${\x}$) from the hidden layer by $g_{\theta'}\left(f_{\theta}(\x)\right)$ .

The unsupervised pre-training tries to minimize the reconstruction cross-entropy loss between $\left\{\x_i \right\}_{i=1}^{n}$ and $\left\{ \tilde{\x}_i \right\}_{i=1}^{n}$ which is defined as follows
\setlength{\arraycolsep}{0.0em}
  \begin{eqnarray}
\mathcal{J}_{\theta,\theta'}(\x;\tilde{\x})& = & \min_{\theta,\theta'}\sum_{i=1}^{n}L\left(\tilde{\x}_i;g_{\theta'}\left(f_{\theta}\left(\x_i\right)\right)\right)
\label{eq:psmmc_rho}
 \end{eqnarray}
with $L\left(\x_i;\z_i\right)$ defined as
  \begin{eqnarray}
L\left(\x_i;\z_i\right) = -\sum_{d=1}^{D}\left( x_{i,d}\log z_{i,d} + (1-x_{i,d})\log (1-z_{i,d}) \right)\nonumber
\label{eq:psmmc_rho2}
 \end{eqnarray}
 where $\z_i$ is short for $g_{\theta'}\left(f_{\theta}\left(\x_i\right)\right)$.
 Problem (\ref{eq:psmmc_rho}) can be solved locally by the well-known back-propagation algorithm.

 When we try to pre-train the $L$-th module with $L>1$ (i.e. the module is not the lowest one), we should first construct its input layer $\x^{(L-1)}$ by transferring $\x^{(0)}$ through the pre-trained networks as follows
   \begin{eqnarray}
\x^{(L-1)} = f_{\theta^{(L-1)}}\left(\ldots f_{\theta^{(l)}}\left(\ldots f_{\theta^{(2)}}\left( f_{\theta^{(1)}}\left( \x^{(0)} \right) \right) \right)\right)
\label{eq:psmmc_rho3}
 \end{eqnarray}
 where $l$ denotes the $l$-th hidden layer (i.e. the $l$-th layer-wise module from the bottom-up), and $\x^{(0)}$ is the original feature vector.

 Here comes the question. What should $\x^{(L-1)}$ reconstruct? Here, we propose to pre-train a clean-speech to clean-speech deep network that accompanies with the noisy-signal to clean-signal deep network, so that we can get $\tilde{\x}^{(l-1)}$ by
    \begin{eqnarray}
\tilde{\x}^{(L-1)} = f_{\tilde{\theta}^{(L-1)}}\left(\ldots f_{\tilde{\theta}^{(l)}}\left(\ldots f_{\tilde{\theta}^{(2)}}\left( f_{\tilde{\theta}^{(1)}}\left( \tilde{\x}^{(0)} \right) \right) \right)\right)
\label{eq:psmmc_rho4}
 \end{eqnarray}
 There are two ways to pre-train the accompanying deep network $\{f_{\tilde{\theta}^{(l)}}\}_{l=1}^{L-1}$ (i.e. the deep neural network for the clean-speech-to-clean-speech reconstruction) in the layer-wise greedy training mode. The first one is to minimize the reconstruction cross-entropy loss via (\ref{eq:psmmc_rho}) with $\tilde{\x}$ as both the input and the target of the module. Another way is to maximize the \textit{logrithmic likelihood} of $\tilde{\x}$ by the efficient \textit{contrastive divergence} algorithm proposed in DBN \cite{carreira2005contrastive}. In this paper, we adopt the former for simplicity.
 Note that we cannot use $\x^{(L-1)}$ to recover $\tilde{\x}^{(0)} $ directly for saving the computation load of constructing $\tilde{\x}^{(L-1)}$, since it's unlikely to describe the extraction network $\left\{f_{{\theta}^{(l)}} \right\}_{l=1}^{L-1}$ of the noisy speech simply by a single hidden-layer reconstruction network $g_{{\theta'}^{(1)}}$.

 In this paper, all activation functions $f_{\theta^{(l)}}\left(\x^{(l-1)}\right)$ and $g_{{\theta'}^{(l)}}\left({\hat{\x}}^{(l-1)}\right)$ are defined as $f_{\theta^{(l)}}\left(\x^{(l-1)}\right)=s\left(\W^{(l)}\x^{(l-1)}+\b^{(l)}\right)$ and $g_{{\theta'}^{(l)}}\left({\hat{\x}}^{(l-1)}\right)=s\left({\W'}^{(l)} {\hat{\x}}^{(l-1)}+{\b'}^{(l)}\right)$ respectively with the function $s(x)$ set to the logistic function $s(x)=1/(1+e^{-x})$ and $\{\W^{(l)},\b^{(l)} \}$ denoted as the weight matrix and the bias term between the $(l-1)$-th and $l$-th layers of the network respectively.

    \begin{algorithm}[t]
    \caption{Denoising-DNN-based VAD.}
    \begin{algorithmic}[1]\label{alg:2}
    \REQUIRE   Feature set $\left\{\x_i^{(0)},\tilde{\x}_i^{(0)},y_i^{(0)} \right\}_{i=1}^{n}$, the depth of the DDNN $L$\\
\ENSURE Feature extraction model $\left\{\theta^{(l)}\right\}_{l=1}^{L}$, and the linear classifier above the model.
\STATE /* Unsupervised denoising layer-wise pre-training */
\FOR{$l=1,\ldots, L$}
\STATE Get $\theta^{(l)}$ by solving $\mathcal{J}_{{\theta}^{(l)},{\theta'}^{(l)}}\left(\x^{(l-1)};\tilde{\x}^{(l-1)}\right)$ defined in equation (\ref{eq:psmmc_rho})
\STATE Calculate $\x^{(l-1)}$ by equation (\ref{eq:psmmc_rho3})
\IF{ $l>1$}
\STATE Get ${\tilde{\theta}}^{(l-1)}$ by solving
 $\mathcal{J}_{{\tilde{\theta}}^{(l-1)},{{\tilde{\theta'}}}^{(l-1)}}\left({\tilde{\x}}^{(l-2)};{\tilde{\x}}^{(l-2)}\right)$ or by the contrastive divergence learning \cite{carreira2005contrastive}.
\STATE Calculate $\tilde{\x}^{(l-1)}$ by equation (\ref{eq:psmmc_rho4})
\ENDIF
\ENDFOR
\STATE /* Supervised fine-tuning */
\STATE Construct the classification-DDNN and fine-tune it by the back-propagation algorithm for the minimum classification error mentioned in Section \ref{subsec:fine-tune}.
\end{algorithmic}
\end{algorithm}

\subsection{Supervised Fine-tuning}\label{subsec:fine-tune}
The supervised fine-tuning phase can be divided into three steps. The first step is to construct the feature extraction part of the DDNN by first discarding the function $\{ g_{{\theta'}^{(l)}}\}_{l=1}^{L}$ and the accompanying deep networks $\{f_{\tilde\theta}, g_{\tilde{\theta}'}\}_{l=1}^{L-1}$ and then stacking all pre-trained functions $\{f_{\theta^{(l)}}\}_{l=1}^{L}$ layer by layer as \cite{hinton2006reducing} did. The second step is to add a linear classifier above the feature extraction part so as to formulate the entire DDNN. The third step is to fine-tune DDNN by the common back-propagation algorithm for the minimum classification error (MCE), where the cross-entropy loss is also used as the surrogate relaxation function.
We call the DDNN for MCE as the classification-DDNN.
Note that another usage of DDNN is to only carry out the first step of the classification-DDNN, and then take the extracted denoising features as the input of some independent classifiers, such as SVM. We call the DDNN for extracting denoising features as the reconstruction-DDNN. We only consider the classification-DDNN in this paper.

\section{Motivation and Related Work}\label{ssec:subsubhead}

The proposed algorithm can be viewed as an idea combination of the stacked denoising autoencoder (SDAE) \cite{vincent2008extracting,vincent2010stacked} and speech enhancement techniques \cite{ephraim1984speech}. SDAE, proposed by Vincent \textit{et al.} in 2008 \cite{vincent2008extracting,vincent2010stacked}, is a novel deep learning technique that has shown comparable performance with DBN. It first adds noise to the original clean features and then takes the noisy features as the input of the module that is to be pre-trained. But it does not try to reconstruct the noisy features. Instead, it tries to recover the original clean features by minimizing the cross-entropy loss or the squared error loss between the reconstructed features and the original clean features. Compared with SDAE, DDNN also tries to recover the clean features, but the noise injected to the clean features is from the real environment instead of from artificial addition.

Speech enhancement techniques, such as the minimum mean square error estimation \cite{ephraim1984speech}, try to estimate the amplitude of the clean speech from the noisy speech observation, which is also known as the \textit{a priori} signal-to-noise ratio (SNR) estimation. The speech enhancement techniques have been widely employed in the VAD research, such as the well-known Sohn VAD \cite{sohn1999statistical}. Compared with the speech enhancement techniques, we construct a deep architecture in a machine-learning perspective for the clean speech estimation with an assumption that the training data has its corresponding clean speech target, while some speech-enhancement-based VADs assume that the background noise is relatively stationary, so that they can trust the statistical parameters updated in the silence period for the clean speech estimation when the speech activity appears. We have to note that many speech enhancement techniques do not need the silence period for the noise spectrum estimation, such as  \cite{cohen2003noise}.

\section{Experiments}\label{sec:analysis}

Seven noisy test corpora of AURORA2 \cite{pearce2000aurora} are used for performance analysis. Four signal-to-noise ratio (SNR) levels of the audio signals are selected, which are $[-5, 0,5, 10]$dB respectively. Each test corpus of AURORA2 contains 1001 utterances, which are split randomly into three groups for training, developing and test respectively. Each training set and development set consist of 300 utterances respectively. Each test set consists of 401 utterances. Note that the corpora in the same background noise scenario but at different SNR levels are split with the same random seed, and have the same manual labels.
  We concatenate all short utterances in each data set to a long one so as to simulate the real-world application environment of VAD. Eventually, the length of each long utterance is in a range of (450,750)s long with the percentages of speech ranging from 54.57\% to 73.32\%.

  The sampling rate is 8kHz. We set the frame length to 25ms long with a frame-shift of 10ms.
We extract 10 acoustic features from each observation. The detailed information of the features are listed in Table \ref{tab:feature}. All features are normalized into the range of $[0,1]$ in dimension.

 \begin{table} [htb]
\caption{\label{tab:feature} {Features and their attributes. The subscript of each feature is the window length of the feature \cite{ramirez2005statistical}.
}}
\centerline{
\scalebox{0.78}{
\begin{tabular}{|l|c|c||l|c|c|}
 \hline
\textbf{ID} & \textbf{Feature} & \textbf{Dimension} & \textbf{ID}&	\textbf{Feature}& \textbf{Dimension}\\
 \hline
1 & Pitch &  1   &		7 & MFCC$_{16}$&20	\\
\hline
2 & DFT & 16 &8&LPC	&12	\\
 \hline
3& DFT$_8$ &  16  & 9&	RASTA-PLP	&17	\\
 \hline
4& DFT$_{16}$&16	&10	&AMS	&	135\\
 \hline
5 & MFCC&  20  &	& \textbf{Total} &273	\\
 \hline
6 & MFCC$_{8}$&  20   &	& &\\
 \hline
\end{tabular}}}
\end{table}

\begin{table*} [thb]
\caption{\label{tab:time} {Accuracy comparison in the \textsf{babble}, \textsf{car}, \textsf{restaurant}, and \textsf{street} noises. The subscripts of the DBN and DDNN are the depths (i.e. the numbers of the hidden layers) of the deep neural networks.}}
\centerline{
\scalebox{0.70}{
\begin{tabular}{|l|c|c|c|c|c|c|c|c|c|c|c|c|c|c|c|c|}
\hline
& \multicolumn{4}{c|}{\textsf{Babble}} & \multicolumn{4}{c|}{\textsf{Car}} & \multicolumn{4}{c|}{\textsf{Restaurant}} & \multicolumn{4}{c|}{\textsf{Street}}\\
 \hline
 & -5dB& 0dB & 5dB & 10dB   & -5dB& 0dB & 5dB & 10dB    & -5dB& 0dB & 5dB & 10dB    & -5dB& 0dB & 5dB & 10dB\\
 \hline
 SVM&   54.61 & 64.46 & 75.97 & 79.53 & 72.20 & 81.59 & 86.34 & 87.60 & 69.04 & 74.22 & 82.09 & 84.83 & 58.32 & 67.98 & 74.88 & 78.12\\
 \hline
 MKSVM& 55.43 & 65.02 & 76.17 & 80.18 & 75.01 & 83.50 & 86.38 & 87.94 & 70.44 & 75.71 & 83.25 & 86.30 & 63.38 & 73.35 & 77.60 & 79.10       \\
 \hline
 \hline
 DBN$_1$&  \textbf{61.03} & 69.01 & 78.83 & 80.99 & 77.24 & 84.10 &  \textbf{87.18} &  \textbf{88.48} &  \textbf{70.23} & \textbf{75.73} & 83.43 &  \textbf{86.12} & 66.63 & 73.15 & 78.47 & 80.42      \\
 \hline
 DBN$_2$&60.81 & 69.24 & 78.94 &  \textbf{81.23} &  \textbf{77.88} &  \textbf{84.14} & 87.04 & 88.44 & 70.10 & 75.68 & \textbf{83.59} & 86.08 &  \textbf{67.41} &  \textbf{73.76} & 78.70 &  \textbf{80.86}      \\
 \hline
 DBN$_3$& 60.55 &  \textbf{69.38} &  \textbf{79.03} & 80.78 & 77.75 & 83.97 & 87.00 & 88.14 & 69.75 & 75.57 & 83.54 & 85.92 & 67.33 & 72.83 &  \textbf{79.03} & 80.49      \\
 \hline
 \hline
 DDNN$_1$&  \textbf{60.69} & 69.42 & 78.61 & 81.39 & 76.06 & 83.86 & 86.77 & 88.17 &  \textbf{69.76} & 75.88 & 83.47 & 86.41 & 66.21 & 72.21 & 79.33 & 81.24    \\
 \hline
 DDNN$_2$& 58.62 & 69.07 & 78.85 & 81.62 & 76.80 & 84.04 & 86.96 & 88.54 & 69.71 & 76.05 &  \textbf{83.90} & 86.62 & 65.51 & 72.72 & 79.17 & 81.53   \\
 \hline
 DDNN$_3$& 57.84 &  \textbf{69.61} &  \textbf{79.14} &  \textbf{81.65} &  \textbf{76.82} &  \textbf{84.22} &  \textbf{87.09} &  \textbf{88.67} & 69.55 &  \textbf{76.04} & 83.78 &  \textbf{86.65} &  \textbf{65.89} &  \textbf{72.82} &  \textbf{79.47} & \textbf{81.71}   \\
 \hline
\end{tabular}}
}
\end{table*}

\begin{table*} [thb]
\caption{\label{tab:time2} {Accuracy comparison in the \textsf{airport}, \textsf{train}, and \textsf{subway} noises. ``{AVR}'' is short for average. ``ALL'' denotes that the AVR is calculated over all noise types and SNR levels. Note that when we calculate the averages, we did not consider the results of the \textsf{babble} noise in $-5$ and $0$ dB, since the manifolds of the speech and background noise are similar in that situation.}}
\centerline{
\scalebox{0.70}{
\begin{tabular}{|l|c|c|c|c|c|c|c|c|c|c|c|c|c|c|c|c||c|}
\hline
& \multicolumn{4}{c|}{\textsf{Airport}} & \multicolumn{4}{c|}{\textsf{Train}} & \multicolumn{4}{c|}{\textsf{Subway}}& \multicolumn{4}{c||}{\textbf{AVR over diff. noise types}} & {\textbf{AVR}}\\
 \hline
 & -5dB& 0dB & 5dB & 10dB   & -5dB& 0dB & 5dB & 10dB    & -5dB& 0dB & 5dB & 10dB   & -5dB& 0dB & 5dB & 10dB  & ALL\\
 \hline
 SVM&   64.48 & 74.26 & 80.94 & 85.21 & 66.24 & 74.29 & 82.91 & 85.28 & 74.75 & 81.24 & 83.58 & 85.18  & 67.51&	75.60&	 80.96&	83.68& 76.93 \\
 \hline
 MKSVM& 65.86 & 75.59 & 82.30 & 85.38 & 68.78 & 76.31 & 83.99 & 85.34 & 79.90 & 84.82 & 86.11 & 87.46   & 70.56&	 78.21&	82.26&	84.53&
 78.89   \\
 \hline
 \hline
 DBN$_1$& 66.18 & 76.63 & 81.89 &  \textbf{86.63} & 68.59 &  \textbf{76.95} &  \textbf{83.65} &  \textbf{85.72} & 78.54 & 82.70 & 85.60 & 85.79  & 71.24&	 78.21&	82.72&	84.88&
 79.26  \\
 \hline
 DBN$_2$& 66.35 &  \textbf{76.66} &  \textbf{81.92} & 86.41 &  \textbf{68.99} &  \textbf{76.95} & 83.49 & 85.68 &  \textbf{79.10} &  \textbf{83.29} & 85.77 &  \textbf{86.25}  &  \textbf{71.64}&	  \textbf{78.41}&	82.78&	 \textbf{84.99}&
  \textbf{79.46}  \\
 \hline
 DBN$_3$&  \textbf{66.62} & 76.38 & 81.85 & 86.50 & 68.89 & 76.14 & 83.56 & 85.62 & 78.95 & 83.26 &  \textbf{85.81} & 86.01  & 71.55&	 78.03&	 \textbf{82.83} &	84.78&
 79.30 \\
 \hline
 \hline
 DDNN$_1$& 66.00 & 76.61 & 82.34 & 86.81 & 68.59 & 77.36 & 83.88 & 85.94 & 77.90 & 83.20 &  \textbf{85.84} &  \textbf{86.64}  & 70.75&	 78.19&	82.89&	85.23&
 79.27\\
 \hline
 DDNN$_2$& 66.80 & 76.86 &  \textbf{82.45} &  \textbf{86.98} & 69.33 & 77.48 & 84.21 & 86.12 & 78.19 & 83.39 & 85.62 & 86.46 & 71.06&	 78.42&	83.02&	85.41&
79.48 \\
 \hline
 DDNN$_3$&  \textbf{67.00} &  \textbf{76.85} & 82.30 & 86.85 &  \textbf{69.44} &  \textbf{77.60} &  \textbf{84.25} &  \textbf{86.16} &  \textbf{78.53} &  \textbf{83.60} & 85.73 & 86.49 &  \textbf{71.21}&	  \textbf{78.52}&	 \textbf{83.11}&	\textbf{85.45}&
  \textbf{79.57}\\
 \hline
\end{tabular}}
}
\end{table*}

The SVM-based VAD, MK-SVM-based VAD, and DBN-based VAD are used for comparison. For the SVM-based VAD, DBN-based VAD, and DDNN-based VAD, we concatenate all 10 features in serial to a long feature vector and take the long feature vector as the input of the classifiers. For the MK-SVM-based VAD, we deal with the features in a similar way with \cite{xu2010simple}.

In respect of the parameter setting, for the SVM-based and MK-SVM-based VADs, the Gaussian RBF kernel is used.
The parameters of SVM and MK-SVM are searched in grid. For the DBN-based and DDNN-based VADs, up to three hidden layers are adopted with the numbers of the hidden units set to $[54,7,7]$ respectively. The learning rate of the unsupervised pre-training is set to 0.004. The maximum epoch of the unsupervised pre-training is set to 200. The learning rate of the supervised fune-tuning is set to 0.005. The maximum epoch of the supervised fune-tuning is set to 130. The batch mode training is adopted. Each batch contains 512 observations. Note that the parameters are selected empirically for a compromise between the training time complexity and the accuracy. We run all experiments 10 times and report the average performances. The reported performance might be further improved by tuning the parameters.

Tables \ref{tab:time} and \ref{tab:time2} list the experimental results. The highlighted contents of each column are the best performance of the referenced DBN-based VAD and that of the DDNN-based VAD on the corresponding noise scenario respectively. From the two tables, we can see that the deep layers of the DDNN-based VADs perform better than the shallower layers, which supports our conjecture in Section \ref{ssec:subsubhead}. Also, the DDNN-based VAD outperforms the SVM-based VAD and the MK-SVM-based VAD. Moreover, the DDNN-based VAD even outperforms the DBN-based VAD in several noise scenarios, which demonstrates its effectiveness. The experimental phenomenon manifested our conjecture in the introduction section about the reason why the deep layers the DBN-based VAD does not outperform the shallow layers. That is, the manifolds of the clean speech and background noise mixed with each other, so that we cannot expect DBN to distinguish the background noise from the noisy speech that contains the manifolds of both the clean speech and the background noise.

\section{Conclusions and Future Work}
\label{sec:conclusion}
In this paper, we have proposed a denoising-deep-neural-networks-based VAD. Specifically, the DDNN training contains two phases. The first phase is to pre-train a deep neural network in an unsupervised denoising greedy layer-wise mode. The second phase is to fine-tune the whole deep neural network as usual. The denoising pre-training makes the DDNN discover the manifold of the clean speech without suffering severely from the disruption of the background noise. Experimental results have shown that the deep layers of the DDNN-based VAD are much more powerful than the shallower layers, and moreover, the DDNN-based VAD outperforms the DBN-based VAD in several noise scenarios.

However, to train a DDNN model, the noisy speech training corpus needs its corresponding clean corpus, which is an ideal situation. Therefore, how to relax this constraint is what we focus on in the future work. Moreover, the experiments are limited to the matching environments, how to make the DDNN-based VAD perform steadily in unmatching environments is another key problem we want to address.

\textbf{Acknowledgment:}
The authors would like to thank the anonymous referees for their valuable advice, which greatly improved the quality of this paper.


%
%
%



\bibliographystyle{IEEEbib}
\bibliography{zxlrefs}

\end{document}